\documentclass[10pt,twocolumn,letterpaper]{article}

\usepackage{cvpr}
\usepackage[utf8]{inputenc} % allow utf-8 input
\usepackage[T1]{fontenc}    % use 8-bit T1 fonts
\usepackage{url}            % simple URL typesetting
\usepackage{booktabs}       % professional-quality tables
\usepackage{amsfonts}       % blackboard math symbols
\usepackage{nicefrac}       % compact symbols for 1/2, etc.
\usepackage{microtype}      % microtypography
\usepackage{graphicx}
\usepackage{amsmath}
\usepackage{amssymb}
\usepackage{subfigure}
\usepackage{epsfig}
\usepackage{verbatim}
\usepackage{multirow}
\usepackage{color}
\usepackage{hhline}
\usepackage{lmodern}
\usepackage[pagebackref=true,breaklinks=true,letterpaper=true,colorlinks,bookmarks=false]{hyperref}

\cvprfinalcopy

\ifcvprfinal\fi
\begin{document}

%%%%%%%%% TITLE
\title{Assessment of Faster R-CNN in Man-Machine collaborative search}

\author{
  Arturo Deza\\
  \small{Dynamical Neuroscience}\\
  \small{UC Santa Barbara}\\
  \texttt{deza@dyns.ucsb.edu} \\
  \and
  Amit Surana \\
  \small{United Technologies Corporation}\\
  \small{United Technologies Research Center} \\
  \texttt{suranaa@utrc.utc.com}\\
  \and
  Miguel P. Eckstein \\
  \small{Psychological and Brain Sciences}\\
  \small{UC Santa Barbara}\\
  \texttt{eckstein@psych.ucsb.edu}\\
}

\maketitle

\begin{abstract}
With the advent of modern expert systems driven by deep learning that supplement human experts 
(e.g. radiologists, dermatologists, surveillance scanners), we analyze how and when do such expert systems enhance human performance
in a fine-grained small target visual search task. 
We set up a 2 session factorial experimental design in which humans visually search for a target  with and without a Deep Learning (DL) expert system. We evaluate human changes of target detection performance and eye-movements in the presence of the DL system.
We find that performance improvements with the DL system (computed via a Faster R-CNN with a VGG16) interacts with observer's perceptual abilities (e.g., sensitivity).
The main results include:
1) The DL system reduces the False Alarm rate per Image on average across observer groups of both high/low sensitivity;
2) \textit{Only}
human observers with high sensitivity perform better than the DL system, while the low sensitivity group does not surpass individual DL system performance, even when aided with the DL system itself;
3) Increases in number of trials and decrease in viewing time were mainly driven by the DL system only for the low sensitivity group.
4) The DL system aids the human observer to fixate at a target by the 3rd fixation, potentially explaining boosts
in performance.
 These results provide insights of the benefits and limitations of deep learning systems that are collaborative or competitive with humans.
\end{abstract}

\section{Introduction}
Visual search is an ubiquitous activity that humans engage in every day for a multitude of tasks. 
Some of these search scenarios are explicit such as: searching for our keys on our desk; while other are
implicit such as looking for pedestrians on the street while driving~\cite{eckstein2011visual}. Visual search may 
also be trivial as in the previous example or
may require stronger degrees of expertise accumulated even over many years 
such as radiologists searching for tumours in mammograms, as well as military surveillance operators, or TSA agents
who must go over a high collection of images in the shortest amount of time. Indeed the successes of Deep Learning Systems have already been 
shown to compete with Dermatologists in~\cite{esteva2017dermatologist} as well as Radiologists~\cite{rajpurkar2017chexnet} 
for cancerous tumor detections. 

Most of the expert systems work has been explored in the medical imaging domain, more specifically in radiology.
Litjens~\textit{et~al.}~\cite{litjens2017survey} compiled an overview of 
300 Deep Learning papers applied to medical imaging. 
In the work of
Kooi~\textit{et~al.}, CNN's and other Computer Aided Detection and Diagnosis (CAD) classifiers 
are compared to each other as automatic diagnosis agents~\cite{kooi2017large}. They find that deep learning systems
rival expert radiologists, as is the recent paper of Rajpurkar~\textit{et~al.} when having radiologists diagnosing pneumonia~\cite{rajpurkar2017chexnet}.
Arevalo~\textit{et~al.} benchmark CNN's to classical computer vision models such as HOG and explore the learned representations
by such deep networks in the first convolutional layer~\cite{arevalo2016representation}.
The majority of studies have evaluated automated intelligent 
agents via classical computer vision or 
end-to-end deep learning
architectures \textit{v.s.} humans. 
See Litjens~\textit{et~al.}~\cite{litjens2017survey} for an overview of 
300 Deep Learning papers applied to medical imaging. 

Other bodies of work regarding collaborative human-machine scenarios in computer vision tasks include: 
image annotation~\cite{russakovsky2015best}, machine teaching~\cite{simard2017machine,JohnsCVPR2015}, visual conversational agents
~\cite{chattopadhyay2017evaluating}, cognitive optimization~\cite{deza2017attention}, and fined-grained categorization~\cite{branson2014ignorant}. 
Conversely, there has also been a recent trend comparing humans against machines in certain tasks with the goal of
finding potential biological constraints that are missing in deep networks. 
These comparisons have been done in object 
recognition~\cite{geirhos2017comparing,eckstein2017humans,pramod2016computational}, perceptual discrimination~\cite{elsayed2018adversarial} and visual attention~\cite{das2017human}.

\begin{figure}[t]
\centering
\includegraphics[width=1.0\columnwidth,clip=true,draft=false,]{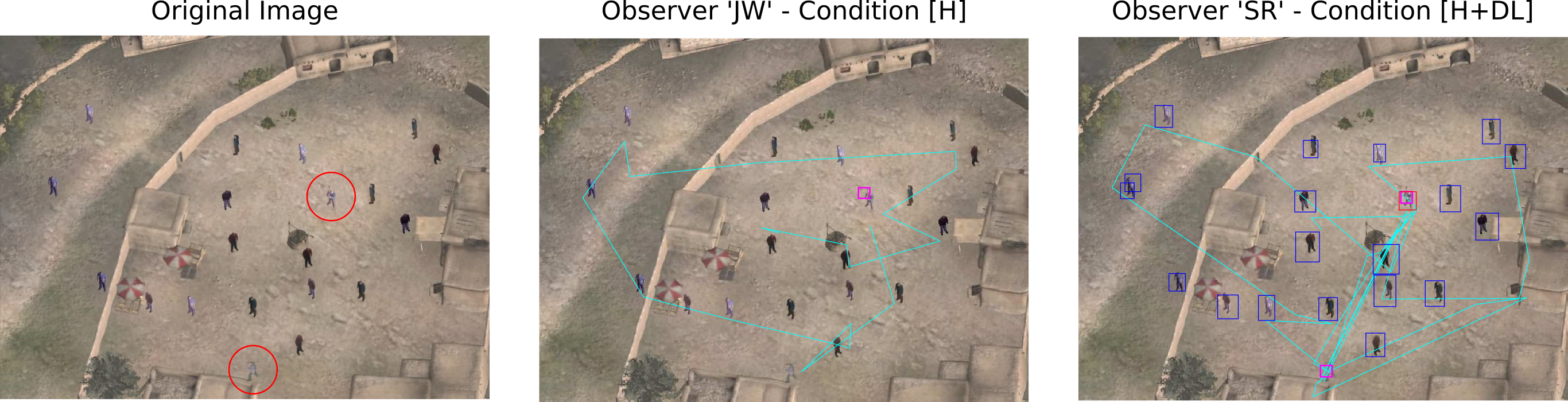}
\caption{An evaluation of potential DL Benefits. Left: The original image with targets circled in red. Middle: Boxes in Magenta are clicks that
observers did on target location. Right: Boxes in blue represent non-target detections and boxes in red represent target detections of the DL System. 
Middle and Right: Saccadic gaze pattern is plotted in cyan.}
\label{fig:System_Benefits}
\end{figure}

In many applications, mixed DL and human teams are a likely next step prior to replacement of the human expert by the expert system~\cite{kneusel2017improving,esteva2017dermatologist,deza2017attention,wolfe2017more,peters2015human}.
Given current paradigms in computer vision technology that rely on bounding box candidate regions proposals and 
evaluations of multiple regions of interest~\cite{malik2016three} as is the case of models from HOG~\cite{dalal2005histograms} and 
DPM~\cite{felzenszwalb2010object} to Faster R-CNN~\cite{ren2015faster} and YOLO~\cite{redmon2015you}, how well do they integrate 
with humans whose visual search system is foveated by nature~\cite{deza2016can,cheung2016emergence,10.1371/journal.pcbi.1005743}? 
We are interested in evaluating the influences of DL systems on human behavior \textit{working together} during visual search for a small \  target in naturalistic scenes
 (see Figure~\ref{fig:System_Benefits}). 

%Previous work regarding integrating eye-movements and computer vision research has usually been related to using deep learning models
%to predict saliency patterns or fixation heat maps. However most of the work in this area has relied on mapping an image to a pattern of eye-movements
%independent on the task. In other words, most computer vision systems have focused on bottom-up saliency prediction, which unfortunately is not 
%representative of real eye-movements in an image given that an image may have different fixation heat maps depending on the task~\cite{koehler2014saliency}.
%We are not the first to bring awareness to this issue, and while saliency prediction is not the focus of this paper, we would like to establish the concept
%of visual search that relies on viewing an image while finding a target which can range from anything such as a gabor patch embedded in noise, 
%to an a person in a scene, a suspiscious item in an X-ray scan, or a tumor in a mammogram.

Perhaps the most relevant work of human-machine collaboration to ours is that of
Kneusel~\&~Mozer~\cite{kneusel2017improving}. 
Such thorough study investigates the influence on human performance of the \textit{visualization} 
of the intelligent 
system's cues used to indicate the likely target locations. %In particular, 
target presence. %
Our main contribution is complementary: 1) We argue for an interaction between the 
human’s observer performance level and that of the intelligent system in determining its 
influence on decisions; 2) We present eye tracking analysis to evaluate the influence of 
the Faster R-CNN on fixation strategies and types of errors:  target not fixated (fixation errors) 
vs. targets fixated and missed (recognition errors). 

In this paper we focus on these questions as there is still ongoing debate in the field regarding
the use of expert Deep Learning systems supplementing human experts.

\section{Overview of Main Experiment}
To analyze how man and machine work together in a visual search task, we designed an experiment with 2 main conditions: Human [H], 
and Human + Deep Learning [H+DL].  The search task was  to find individuals holding weapons among groups of individuals without weapons.  The people were embedded in a complex scene.
In the following sub-sections, we describe in detail the experiments  (stimuli, experimental design \& apparatus). We evaluated the influence of the Faster-RCNN on the following human behavioral measures during visual search:

\begin{enumerate}
 \item Target detection performance.% (Hit Rate, False Alarms per Image, Misses, Receiver Operating Characteristic)
 \item Receiver Operating Characteristic (ROC) curves.
 \item Viewing time and number of trials.%
 \item Pattern of eye movements. 
\end{enumerate}

\subsection{Creation of Stimuli}
\begin{figure*}[!t]
\centering
\includegraphics[width=2.0\columnwidth,clip=true,draft=false,]{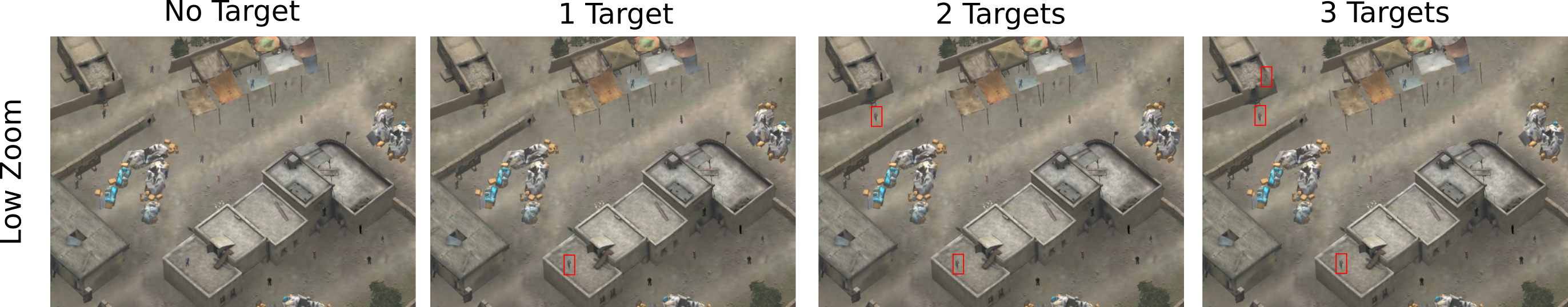}
\caption{An example of a family of stimuli used in our experiment with the same image rendered with different number of 
targets (from left to right). The figure is better viewed when zoomed in, and illustrates the difficulty of visual search. 
Targets are individuals holding weapons, and they 
have been highlighted in red for visualization purposes.}
\label{fig:SCORCH_CV_Stimuli}
\end{figure*}

We selected 120 base images with no targets from the dataset of Deza~\textit{et~al.}~\cite{deza2017attention} that contained a variety of rendered outdoor scenes with different levels of clutter and three levels of zoom. We then randomly picked 20 locations (uniformly distributed) within each image to locate targets (individuals with weapon) and distractors (individuals without weapons).  
We ran a canny edge detection~\cite{canny1987computational} filter to compute major edges in each images such as walls, trees and other structures. 
If one of the previously randomly selected
locations landed on an edge, we would resample uniformly from any place in the image until an edge-less location was found. Our image generation model would also 
re-sample a candidate location
if they were overlapping with a previous person location. 
Once the 20 locations were verified, we generated 4 different versions of the same background image such that
each version had $k=\{0,1,2,3\}$ targets (totalling $4\times120$) with the rest of candidate locations having non-targets (\textit{a.k.a.} 
friends or persons without weapons).  We used Poisson blending~\cite{perez2003poisson} on each of the locations %generated by our sampling generator
to blend the inserted individuals into the background scene.
%All candidate targets used in this rendering process have been previously annotated from the SCORCH dataset.
Each image was rendered at $1024\times760$ px.
%All rendered images are stored in computer memory for so no real-time generation is done during the actual
%experiment.
Example scenes of the Low Zoom condition can be seen in Figure~\ref{fig:SCORCH_CV_Stimuli}, 
where the difficulty of trying to find a target (a person with a weapon) is quite high.

\subsection{Experimental Design}
Our main experiment had a $2\times2$ factorial design to dissociate improvements caused by the DL System and those due to human learning. 
In the experimental design each observer participated in two consecutive sessions in one of the following orders: [H,H] (Human, Human),
[H,H+DL] (Human, Human + Deep Learning), 
[H+DL,H] (Human + Deep Learning, Human) 
and [H+DL,H+DL] (Human + Deep Learning, Human + Deep Learning). 
Comparison of performance improvements in the Human, Human + Deep Learning vs. the Human, Human conditions allows determining whether performance increases are due to the DL system or simply \textit{human learning} effects. 
In addition, we are interested in dissecting learning and ordering effects as it could be the case
that the performance differences in the second session are independent of the use of the DL system.

To make a direct comparison between the DL System and humans, the observers reported the number of individuals with weapons (targets). Observers also spatially localized the targets by  clicking on the location of the detected target individuals on a subsequently presented image that contained the background image and bounding box locations (but no individuals) of all the potential
target candidates. This evaluation paradigm is well matched to the DL system which also localizes targets with no apriori knowledge of how many targets are present in an image.  The number of target per images was randomly selected with a truncated Poisson Distribution where:
\begin{equation}
P_k = P(X=k) = \frac{\alpha^k e^{-\alpha}}{k!} 
\end{equation}
We fixed the value of $\alpha=1$ which represents the average number of targets per trial, such 
that $P_0=0.375$; $P_1=0.375$; $P_2=0.1875$ and
$P_3=0.0625$.

\subsection{Apparatus}
An EyeLink 1000 system (SR Research) was used to collect Eye Tracking data at a frequency of 1000Hz. 
Each participant was at a distance of 76 cm from a LCD screen
on gamma display, so that each pixel subtended a visual angle of $0.022\deg/$px. All images were 
rendered at $1024\times 760$ pixels $(22.5\deg\times 16.7\deg)$.
Eye movements with velocity over $22\deg/s$ and acceleration over $4000\deg/s^2$ were categorized as saccades. 
Every trial began with a center fixation cross, where
each subject had to fixate the cross with a tolerance of $1\deg$. 

\begin{figure*}[!t]
\centering
\subfigure[Condition \text{[H]}: Human Observer. In this condition there is no aid or cueing of targets. At the end of the trial, ground truth person locations 
(colored in black) are overlayed in the image to assist observers on clicking the location of potential targets.]{
    \includegraphics[width=1.0\columnwidth,clip=true,draft=false,]{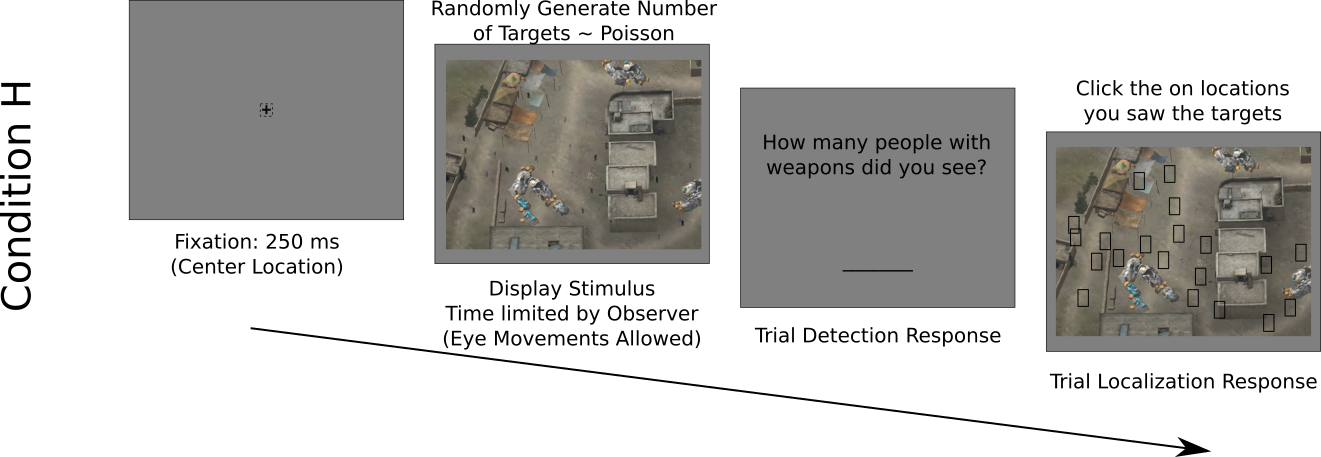}
    \label{fig:Multi_Condition_A}
    \centering
}
\subfigure[Condition \text{[H+DL]}: Human Observer + Deep Learning System. In this condition, candidate targets are cued by the DL system 
with color coded bounding boxes. Colors: Red is a potential foe, and Blue a potential friend.% Notice that the system does not manage to detect all candidate persons in the scene.
]{
    \includegraphics[width=1.0\columnwidth,clip=true,draft=false,]{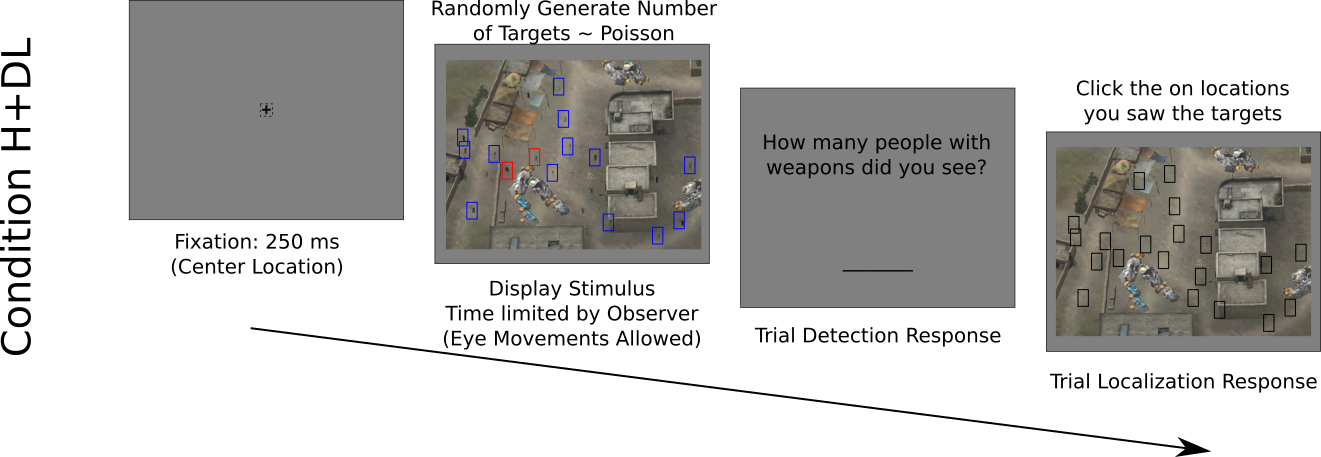}
    \label{fig:Multi_Condition_B}
}
\caption[]{An overview of the 2 conditions tested in the 
multiple target search experiment where we evaluated the 
benefits of a DL System in human visual search as well 
as the possible added benefits in terms of speed, accuracy and eye movements. 
Targets in these images are displayed at $0.45\times 0.90$ d.v.a. 
Data was collected for conditions [H,H]; [H,H+DL]; [H+DL,H]; and [H+DL,H+DL].
}
\label{fig:Multi_Conditions}
\end{figure*}

\section{Training and Testing for Man and Machine}

\subsection{Human: Training and Testing}
A total of 120 observers divided in four groups 
of 30 performed the [H,H], [H,H+DL], [H+DL,H], [H+DL,H+DL] sessions respectively.

\textbf{Training:} Each observer engaged in 3 practice trials at the beginning of each session. Feedback was given at the end
of each practice trial analogous to providing a supervised signal.

\textbf{Testing}: Observers were instructed to optimize two general goals: The first was to \textit{maximize} 
the total number of trials on each of the 20 minute sessions. The second was to \textit{maximize} their performance when engaging in visual search.
We emphasized that they had to do well maximizing both goals, such that they should not rush over the trials and do a poor job, but neither should they over dwell
on search time for every image. No feedback was given at the end of each trial. See Figure~\ref{fig:Multi_Conditions} for experimental flow.
%It was expected that each participants developed a pace of their own, and that potential learning effects carrying over
%from the first session to the second would be detected through the [H,H] condition. 

\subsection{Deep Learning System: Training and Testing}
We trained a Faster R-CNN object detection framework~\cite{ren2015faster} which uses a VGG-Net~\cite{simonyan2014very} for object detection and the 
candidate region proposals. 
%This framework
%was chosen since it is has been shown to be more robust than most other object detectors publicly available~\cite{huang2017speed}.
We picked Faster R-CNN over YOLO~\cite{redmon2015you}, SSD~\cite{liu2016ssd}, R-FCN~\cite{dai2016r} 
given the experiments done by Huang~\textit{et~al.} where they show that Faster-RCNN overperforms the other models
performance-wise~\cite{huang2017speed}.
While running multiple object detectors
in this experiment would have enriched our evaluation, we are limited by the fact that we will need multiple subjects to be ran for each DL system. One of the other reasons we did not pick YOLO over Faster-RCNN is that Real-Time detection in our experiments is not an issue given that we saved all the detected bounding boxes and scores in memory. In addition YOLO might not perform as well as Faster-RCNN
for detecting small objects~\cite{redmon2017yolo9000}.
%We also did not pick YOLO~\cite{redmon2015you} in for our experiments since real-time object detection is 
%not critical in our experiments, and Faster R-CNN is known to perform better than YOLO on other object detection tasks.
Finally, the wide-spread of VGG-Net and Faster-RCNN make them both ideal candidates for our experiments.

\begin{figure}[!h]
\centering
\includegraphics[width=0.9\columnwidth,clip=true,draft=false,]{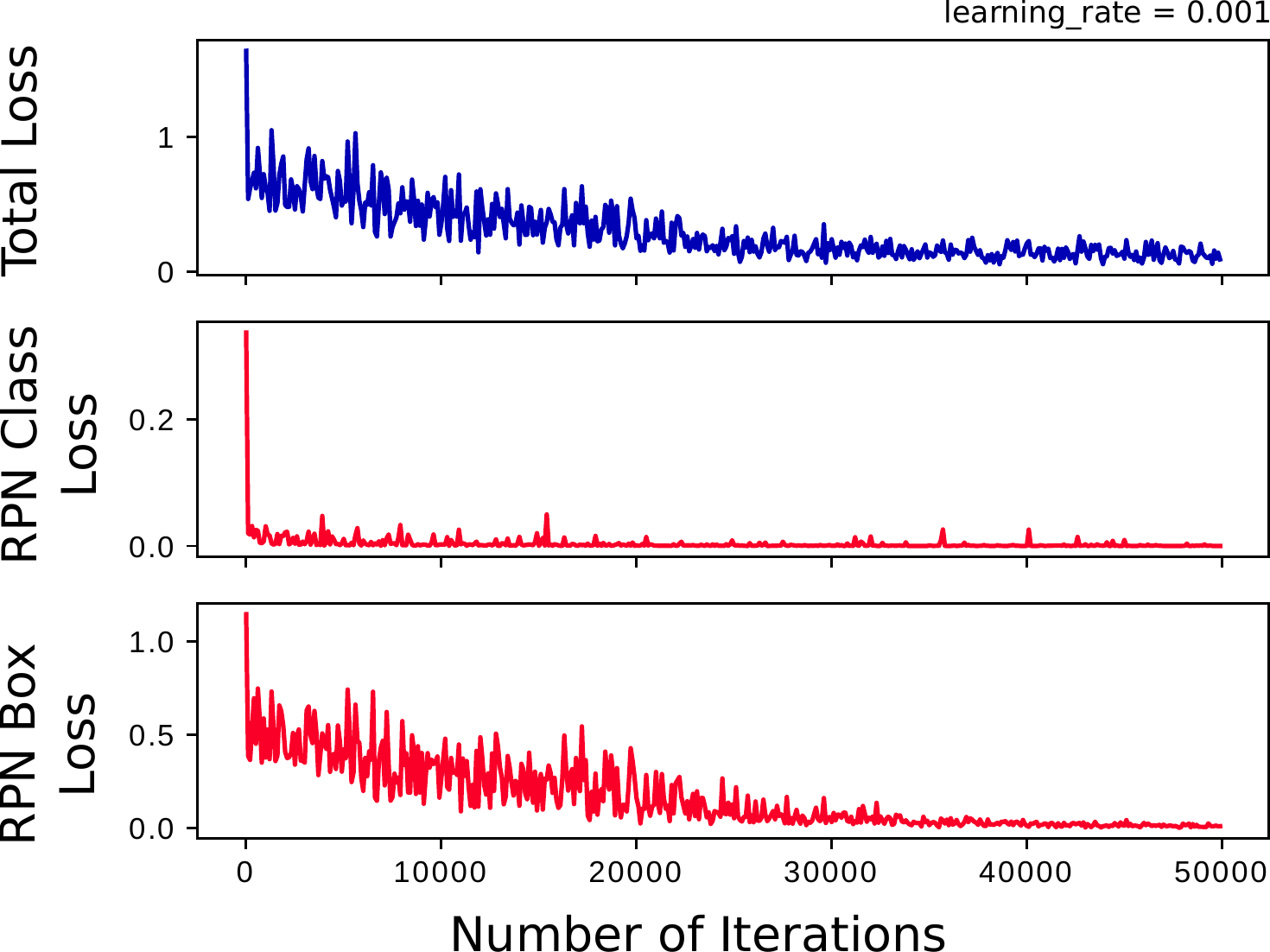}
\caption{Training loss for the Faster-RCNN trained after 50k iterations. We used the model trained after 
5000 iterations to avoid over-fitting. Having a relatively high performing (but not perfect) system is ideal to split observers into high 
and low sensitivity groups for post-hoc analysis.}
\label{fig:Train_Faster_RCNN}
\end{figure}

\textbf{Training}: We trained the network on tensorflow~\cite{abadi2016tensorflow} for over 5000 iterations as shown in Figure~\ref{fig:Train_Faster_RCNN}, after having it pre-trained with 70000 iterations on a collection of images 
from ImageNet achieving standard
recognition performance. 
%We used the publicly available tensorflow code to train our network. 
%The network was trained in tensorflow~\cite{abadi2016tensorflow}.
The images fed to the network for training were $420=7\times20\times3$ images, 
consisting of 7 rotated rotated versions and 20 person inputs (10/10 friends/foes) for each of the 3 target sizes. Small rotations, crops, mirroring and translations
were used for data augmentation.
The images that were rendered for testing had never been seen from the network, and were rendered with a mix of randomly sampled individuals with and without weapons from the held out dataset.

\textbf{Testing}: Candidate bounding boxes developed by the system always overlayed on possible person locations irrespective of whether the individual carried a weapon. Thus the DL 
System never produced a Location-driven False Alarm, all mistakes delivered by the system were recognition/classification based. Bounding box candidates with a threshold
lower than $\eta=0.8$ were discarded, and overlaying bounding boxes (doubles) were removed with non-maximal suppression (NMS).

With these configurations both the 
DL System and the Human are prone to make the same type of judgments and mistakes. For example:
1) Humans are not allowed to click on the same locations more than twice (computer as well given NMS); 2) The Human and DL system 
both have a finite collection
of possible locations from where to select the target locations. 
In addition, the experiment is free-recall for humans as they are allowed to report any number of targets per image without prior information. 
The DL system has the same criteria since the computation of target location via the Region Proposal Network (RPN)
does not depend on any prior of the number of targets seen in the image. 

\section{Results}
\label{Results}
The results shown in this paper focus on the subgroup of trials that showed \textit{small targets} given the greater difficulty in 
detection for both man and machine. 

\textbf{Observer Sensitivity:} We quantified the influence of the DL system across groups of observers with different abilities to find the target (hit rate). We split the participants from the [H,H+DL] condition into two groups
contingent on their \textit{sensitivity} (hit rate): the first group was the high sensitivity group who had a hit rate higher than the DL system
in the first session, conversely the second group was the low sensitivity group who had a lower hit rate than the DL system. We ran an unpaired
t-test to verify that there were indeed performance differences, and found 
a significant difference $t(27)=3.64,p=0.0011$ for the high sensitivity group $(M_H=83.16\pm2.00\%)$ and the low sensitivity 
group $(M_L=65.52\pm4.04\%)$. This effect was visible across all other 
conditions: [H+DL,H] with $t(28)=3.40,p=0.0020$, $(M_H= 89.34\pm2.15\%)$, $(M_L=73.66\pm 3.67\%)$; 
[H,H] with $t(27)=3.96,p<0.001$, $(M_H=85.68\pm2.06\%)$, $(M_L=65.75\pm3.46\%)$; and 
[H+DL,H+DL] with $t(27)=2.21,p=0.0351$, $(M_H=85.24\pm3.68\%)$, $(M_L=71.79\pm2.45\%)$.

\begin{figure*}[t]
\centering
\includegraphics[width=2.0\columnwidth,clip=true,draft=false,]{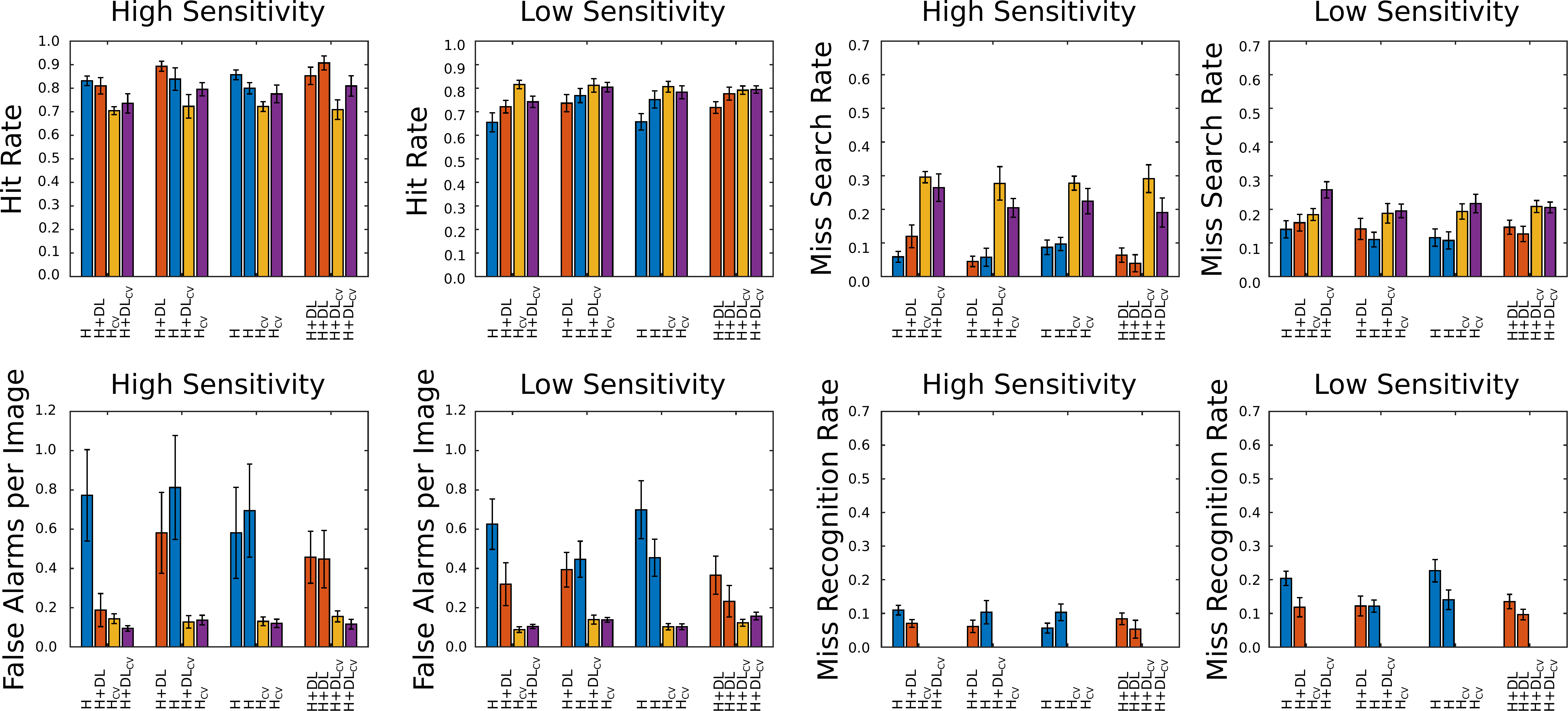}
\caption{Partition of observer performance given by Sensitivity (Hit Rate) higher or lower than the machine. Hit Rate, False Alarms per Image, Miss Search Rate and Miss Recognition Rate are shown for each group. Session color code:
Blue: Human without DL ; Orange: Human with DL ; Ocre: DL on 1st session; Purple: DL on 2nd session.}
\label{fig:Partition1}
\end{figure*}

\subsection{Target Detectability}
\label{sec:Target_Metrics}
In the following subsection we describe the collection of the metrics used in our analysis that come from 
the signal detection theory literature~\cite{green1988signal} and medical imaging/radiology 
(search and recognition errors)~\cite{krupinski2010current}. We group such metrics contingent on the sensitivity
of each observer and plot these values in Figure~\ref{fig:Partition1}.
%To report our final results as seen in Table~\ref{tab:table1}, each of these metrics was 
%averaged across
%all observers. It should also be noted that the number of trials done in each session for each observer may vary.

\begin{enumerate}
\item \textbf{Hit Rate per Image} (HR): The total number of targets correctly selected at divided by the total number of targets in the image.
\item \textbf{False Alarms per Image} (FA): The total number of false positives (disctractor individuals without weapons incorrectly labelled as targets).
\item \textbf{Miss Rate per Image} (MR): 1.0 - Hit Rate per Image. We divide the Miss Rate in two types:
\begin{itemize}
\item \textbf{Search Errors Rate per Image} (SER): The total number of targets that were not foveated and missed divided by the total number of targets in the image. For the machine we consider these as bounding boxes where the output probability did not exceed the confidence threshold $(\eta)$, as one could otherwise argue that the machine `foveates' everywhere.
\item \textbf{Recognition Errors Rate per Image} (RER): The total number of targets that were foveated, yet incorrectly perceived as friends (when they are actually foes) divided
by the total number of targets in the image. 
It should be observed that 
%Recognition Error Rate per Image and the Search Error Rate per Image 
RER and SER 
should add up to the Miss Rate per Image.
\end{itemize}
%\item \textbf{Index of Sensitivity ($d'$)}: It is defined as $d' = Z^{-1}(HR) - Z^{-1}(FA)$, 
%where $Z^{-1}(\cdot)$ is the inverse of the normal cumulative function.
%This quantity is generally of interest when there are an unbalanced number of trials
%of target present and target absent in a yes/no task. Consider the following case scenario: an observer responds ``target present'' on all trials
%of an experiment that has $80\%$ images of target present and $20\%$ images of target absent. 
%The observer has proportion correct $(PC) = 80\%$ (which is misleading: `there are two classes, and (s)he is doing highly above chance'), but $d'=0$, hence no target sensitivity. 
\end{enumerate}

We performed two sets of mixed factor design ANOVA's for within conditions: [H] and [H+DL]; between conditions: 
order effects [H,H+DL] and [H+DL,H]; and between subjects. Each mixed ANOVA was ran separately for the high and low sensitivity groups. We found the following
results:

\textbf{False Alarms per Image}: A main effect of \textit{reduction} 
of False Alarms with the presence of the DL system for both the high and low sensitivity group: %$F(1,228)=13.74,p=0.0003$.
$F_H(1,24) = 7.23,p=0.01$, and $F_L(1,24) = 4.93,p=0.03$.

\textbf{Search Error Rate:} No significant differences in terms of search error rate between conditions. 
Although we did find that on average the search error rate was lower for the high sensitivity group: unpaired, 
two-tailed, $t(116)=-3.633,p<0.0001$.

\textbf{Recognition Error Rate:} No reduction in recognition error rate for the high sensitivity group,
but a marginal main effect for reduction in recognition error rate for the low 
sensitivity group in the presence of the DL system $F_L(1,32)=3.85,p=0.058$,
as well as a marginal ordering effect (showing [H+DL] or [H] first) $F_L(1,32)=3.96,p=0.055$.

%\begin{figure}[!h]
%\centering
%\includegraphics[width=0.9\columnwidth,clip=true,draft=false,]{Summary_Stats_Total.pdf}
%\vspace{-10pt}
%\caption{Summary target detectability metrics for multiple target sizes and across the three experimental conditions.}
%\label{fig:Summary_Stats}
%\end{figure}

% Figure relating to the first saccade to the observer
%\begin{figure}[!h]
%\centering
%\includegraphics[width=0.9\columnwidth,clip=true,draft=false,]{Num_Fixations.pdf}
%\vspace{-10pt}
%\caption{Number of Fixations to first foveate any target for multiple target sizes across our three experimental conditions. Stars indicate statistically significant differences
%with $(p<0.05)$.\arturo{Potentially change color of barplots for A and B?}}
%\label{fig:Num_Fixations}
%\end{figure}

\begin{figure*}[t]
\centering
\includegraphics[width=2.0\columnwidth,clip=true,draft=false,]{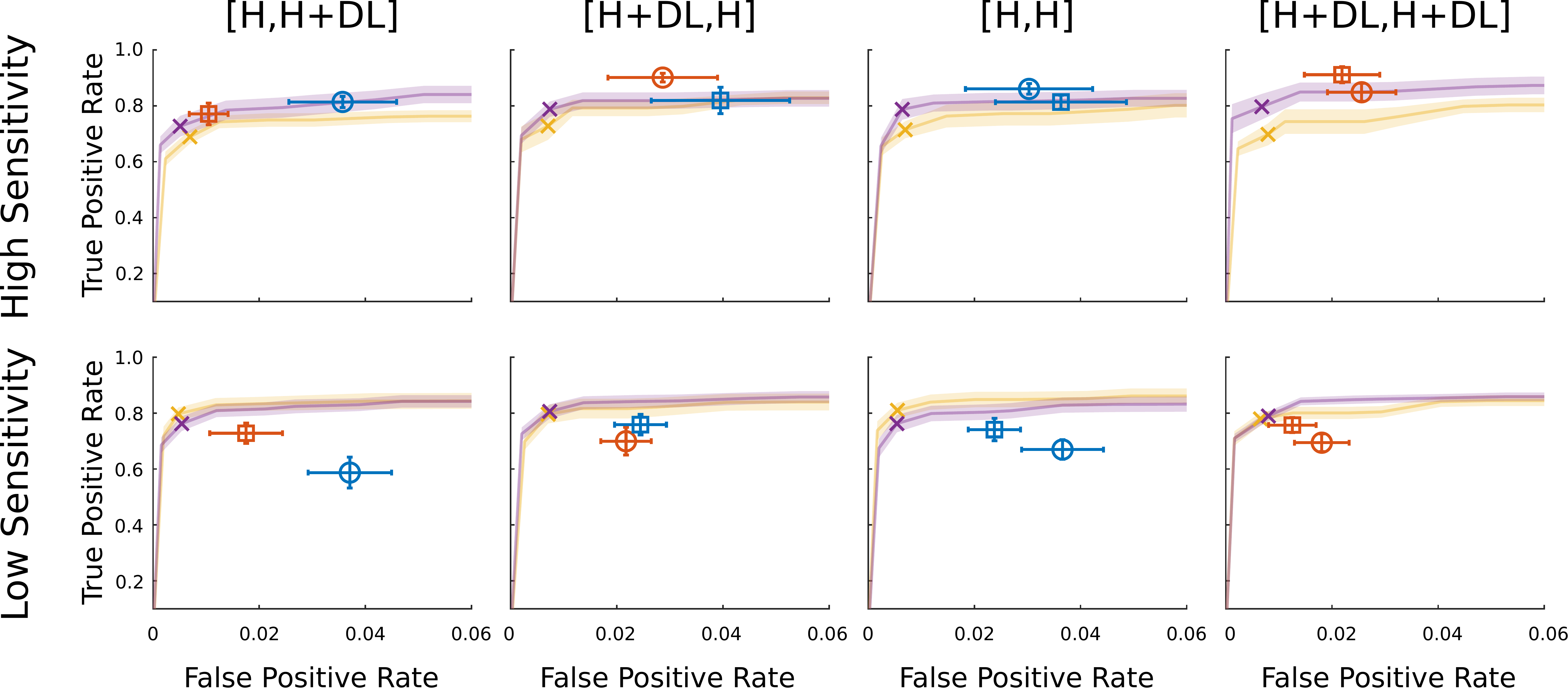}
\caption{ROC plots that compare the performance of the Human and the DL system separately and working collaboratively.
The plots are split by High~/~Low sensitivity, and Experimental Condition: [H,H+DL], [H+DL,H], [H,H] and [H+DL,H+DL].
ROC's in ocre and purple show the performance of the 
DL System independently for the first and second session respectively. The cross indicates the operating point along the curve at $\eta=0.8$. 
For the human observer a circle is the first session,
and a square the second session. Blue and orange indicate presence of the DL system when engaging in visual search.}
\label{fig:ROC_Analysis}
\end{figure*}

\subsection{Assessment of the Human and Machine Receiving Operating Characteristics}

Similar to the work of 
Esteva~\textit{et~al.}~\cite{esteva2017dermatologist}, we decided to investigate how do humans perform compared to the DL system when 
the system performs individually along its entire receiver operating characteristic (ROC) curve, including its operation point 
at $\eta=0.8$. It may be possible that we find that the DL system performs much better overall than the human observers even for the high sensitivity group,
as a higher sensitivity might also imply high false alarm rates and thus less discriminability. This is an effect that can usually be explained
within the context of signal detection theory~\cite{green1988signal}.
If the ROC point of the human observers with or without assistance
is outside of the DL ROC curve (ocre and purple for the each of the 2 sessions respectively), then we can say that the humans observers collectively 
perform better than the machine.

To compute the ROC curve per image we require both the TPR (True Positive Rate) and FPR (False Positive Rate) 
per image $I$. Note that FPR is not be confused with False Alarms per Image 
as plotted in Figure~\ref{fig:Partition1}.
If $h$ is the number of hits the observer performs on the image, and $f$ 
the number of false alarms restricted to the clicked bounding box locations:
We will compute $TPR=h/G$, and $FPR=f/(N-G)$, 
where $N=20$ is the total number of possible bounding boxes that
an observer has to choose from to make a selection for target present, 
and $G$ is the number of true targets there are in the image $(0,1,2~\text{or}~3)$.
These statistics were averaged for both the machine to plot an entire ROC curve, and for the human observers plotting the ROC points 
as depicted in Figure~\ref{fig:ROC_Analysis}.

To analyze variability in the observers behaviour as well as decision strategies we will use estimates of target 
detectability ($d'$) and 
decision bias ($\lambda$) s.t. 
\begin{equation}
d' = \Phi^{-1}(TPR) - \Phi^{-1}(FPR)
\end{equation}
and 
\begin{equation}
\lambda = - \Phi^{-1}(FPR)
\end{equation} 
where $\Phi^{-1}$ is the inverse of the cumulative normal distribution. 

In what follows of the remaining subsection we focus on comparing two types of conditions across each others along previously mentioned metrics. These are mainly:
[H,H+DL] \textit{vs} [H,H], to investigate how the observer ROC changes in the second session with the presence of the DL system, and also
[H+DL,H] \textit{vs} [H+DL,H+DL] which investigates if the observer's  signal detectability and criterion change as a function discarding/continuing 
the DL system in the second session.

%We find that \arturo{complete.}

\textbf{Detectability $(d')$:} We performed an unpaired t-test 
across the second sessions comparing [H,H+DL] \textit{vs} [H,H],
and [H+DL,H] \textit{vs} [H+DL,H+DL], and did not find any statistically significant changes in $d'$.

\textbf{Decision bias $(\lambda)$:} 
Only the high sensitivity group showed differences in bias when the
DL system was removed in the second session $t(24)=2.62,p=0.01$.
$\hat{\lambda}_{H+DL}=2.09\pm0.05$ \textit{vs} $\hat{\lambda}_{H+DL}=1.79\pm0.12$ in the [H,H+DL] \textit{vs} [H,H]
condition.

We finally summarized the detectability and bias scores across all observers, pooled over both sessions, and split by sensitivity and condition [H] vs [H+DL], and compared these to the machine in Table~\ref{table:SCORCH_Summary}:

\begin{table}[h]
\scriptsize
\centering
\begin{tabular}{|c|c|c|c|c|}
 \hline
 & \multicolumn{2}{|c|}{detectability $(d')$} & \multicolumn{2}{|c|}{bias $(\lambda)$} \\
 \hline
 & [H] & [H+DL] & [H] & [H+DL] \\
 \hline
 High & $2.84\pm0.10$ & $3.13\pm0.09$ & $1.82\pm0.05$ & $1.95\pm0.04$ \\
 Low & $2.42\pm0.10$ & $2.62\pm0.08$ & $1.83\pm0.03$ & $2.00\pm0.03$ \\
 \hline
 DL & \multicolumn{2}{|c|}{$2.78\pm0.04$} & \multicolumn{2}{|c|}{$1.96\pm0.02$} \\
 \hline
\end{tabular}
\caption{Human vs DL system performance}
\label{table:SCORCH_Summary}
\end{table}

It is clear that when removing any learning effects of session order, that \textit{only}
human observers with high sensitivity perform better than the DL system, while the low sensitivity group does not surpass individual DL system performance, even when aided with the DL system itself.

\subsection{Analysis of Viewing Time and Number of Trials}
\begin{figure}[h]
\centering
\includegraphics[width=0.9\columnwidth,clip=true,draft=false,]{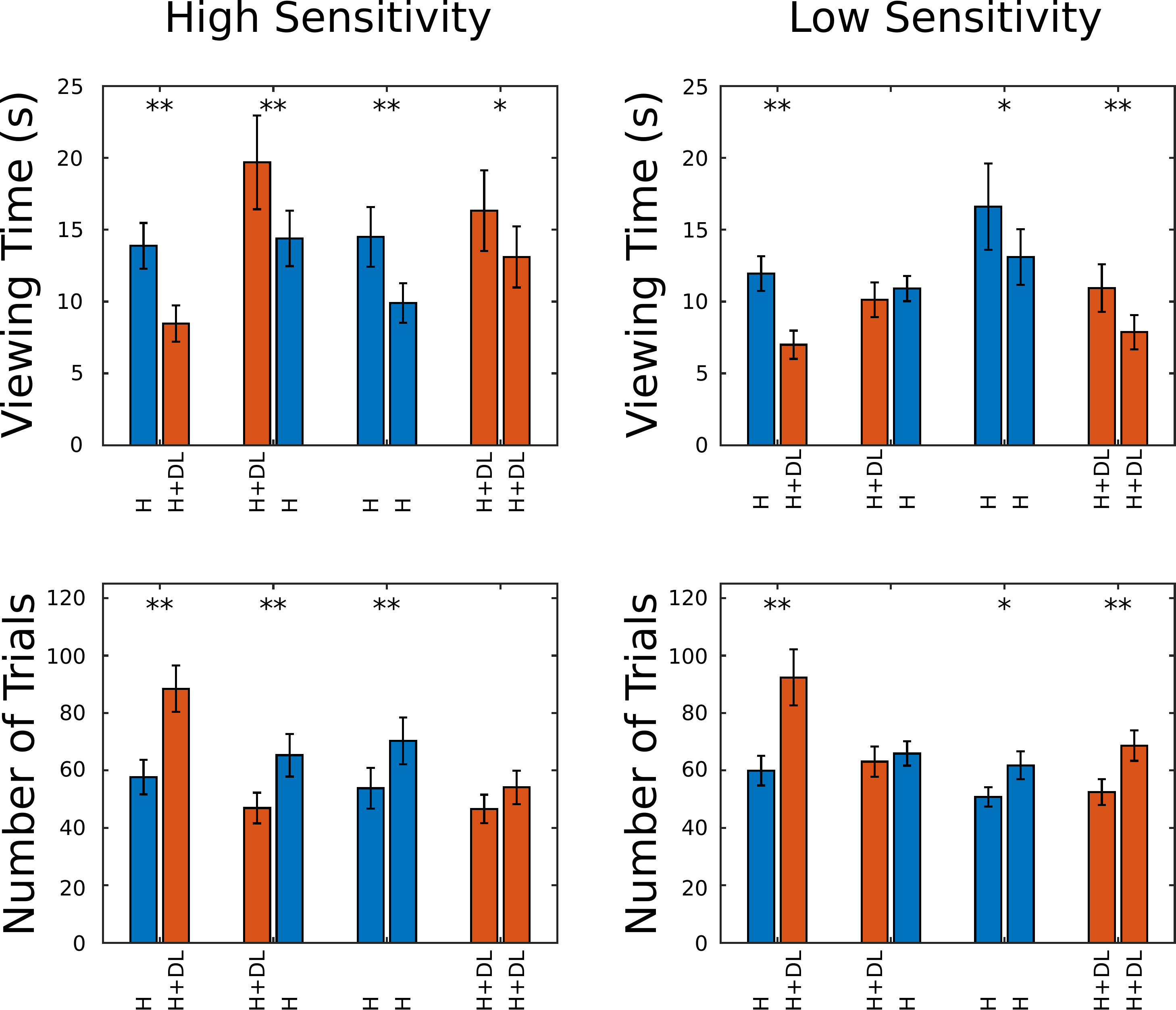}
\caption{Viewing Time and Number of Trials split by high and low sensitivity observers. Blue represents the human observer [H],
and orange represents the Human and Deep Learning system working together [H+DL]. 1 star represents a two-tailed independent t-test with $p<0.05$,
while 2 stars represents $p<0.01$.}
\label{fig:Time_Trials_Split}
\vspace{-10pt}
\end{figure}

\textbf{Viewing Time}: We found significant ordering effects for the high sensitivity group in viewing time spent per 
trial $F(1,24),p=0.05$, but did not find any effects for the presence of the DL system. However, we did find an interaction
for order and presence of the DL system $F(1,24)=24.00,p<0.0001$. As for the low sensitivity group we did not find an ordering 
effect $F(1,32)=0.74,p=0.40$, and rather did find a main effect in the presence of the DL system $F(1,32)=10.56,p=0.003$. This effect
is shown in Figure~\ref{fig:Time_Trials_Split} as a decrease in viewing time. In addition we found an interaction of order and presence of the DL system
$F(1,32)=5.6,p=0.02$. 

Perhaps a striking and counter-intuitive difference 
worth emphasizing is that the low sensitivity group spends \textit{less} time than the high sensitivity group viewing each image
when the system is on independent of order. Although this is understandable as our splits are driven by the performance of the observer on their first session
independent of the presence of the DL system or not. In general, bad performing observers will very likely go over the image faster than high performing observers
which are more careful when examining the image. Indeed, 
to account for differences in the splits, we ran an unpooled t-test to compare between all the [H+DL] 
sessions in the high and low sensitivity groups (across all orders) and found that 
the average viewing time (VT) differences were $VT_H = 14.35\pm1.37$ seconds, and $VT_L= 9.05\pm0.67$ seconds, with $t(117)=3.84,p<0.0001$.

%The average viewing time decreases for the high sensitivity group $F(1,226)=7.61,p=0.0063$, and the the average number of trials
%also decreases after on the second session $F(1,226)=14.72,p=0.0002$. We did not find a savings in viewing time contingent on the 
%DL system $F(1,226)=2.7,p=0.1019$,
%% which shows a start contrast with the work of Deza~\textit{et~al.}~\cite{deza2017attention}, which shows an acceleration on number of trials 
%while keeping the subject target detection metrics constant with the presence of a Cognitive Optimizer. We do however find an interaction
%between the sensitivity group and the use of the DL system $F(1,226)=7.07,p=0.0084$, where observers spend about the same amount of time
%scanning for targets without the DL system for any sensitivity group, while the DL system makes the high sensitivity spend more viewing time.

\textbf{Number of Trials:} All the results we found for Viewing Time are analogous and statistically significant when analyzing number of trials -- 
as the total time per session in the experiment is constricted
to 20 minutes, and both these quantities are inversely proportional to each other. Figure~\ref{fig:Time_Trials_Split} shows such equivalence and 
how a low viewing time generally translates to a high number of trials across all conditions.

\subsection{Analysis of Eye-Movements}

\begin{figure}[t]
\centering
\includegraphics[width=1.0\columnwidth,clip=true,draft=false,]{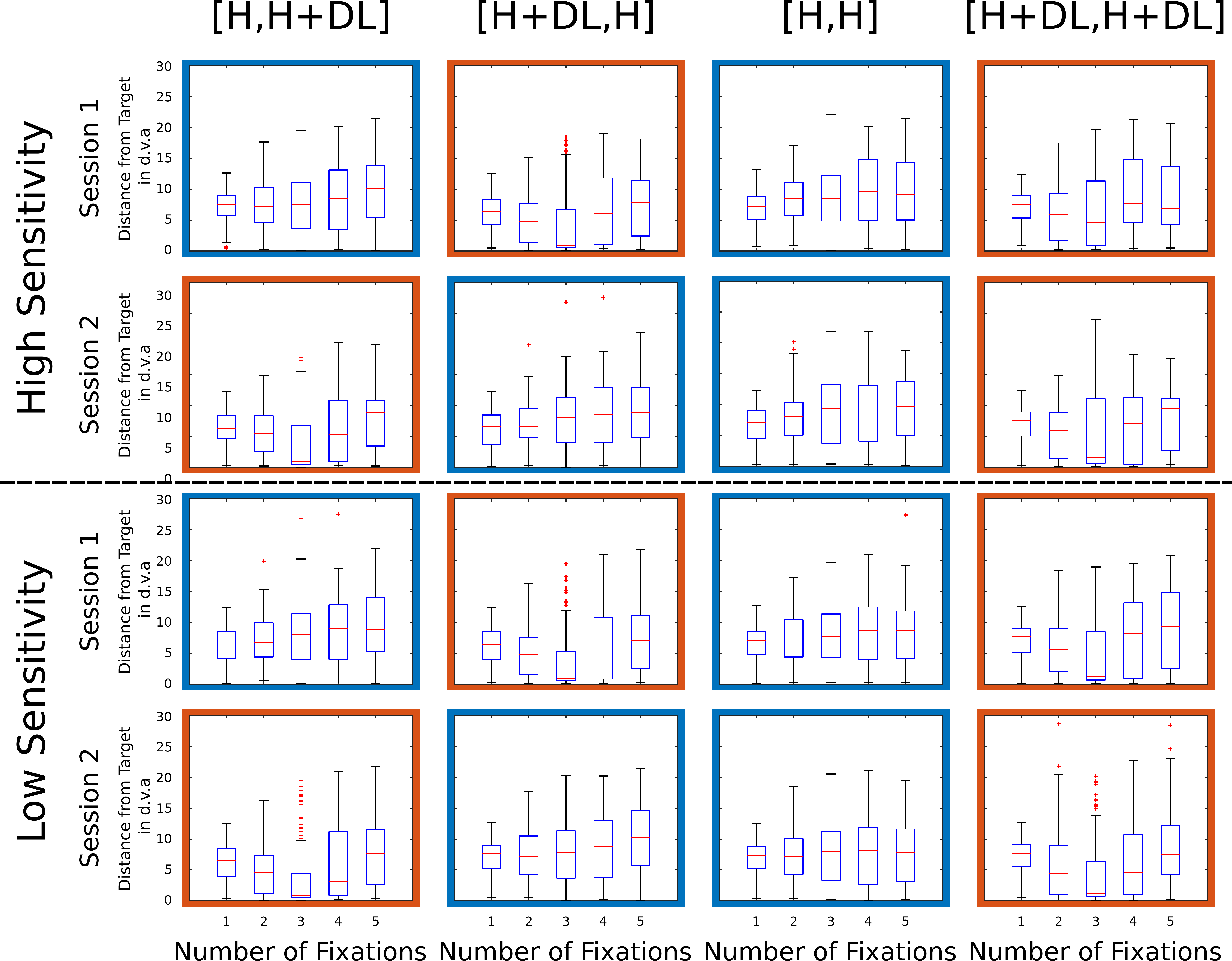}
\caption{Boxplots of the fixation distance to the 
first target foveated in degrees of visual angle (d.v.a). The 
Expert System aids the human by assisting him/her to fixate the target at $\sim1\deg$ by the 3rd fixation (orange barplots). 
This visual search strategy is only present when the
Expert System is on -- independent of the session order.}
\label{fig:FixationDistance}
\end{figure}

\begin{figure}[!t]
\centering
\includegraphics[width=1.0\columnwidth,clip=true,draft=false,]{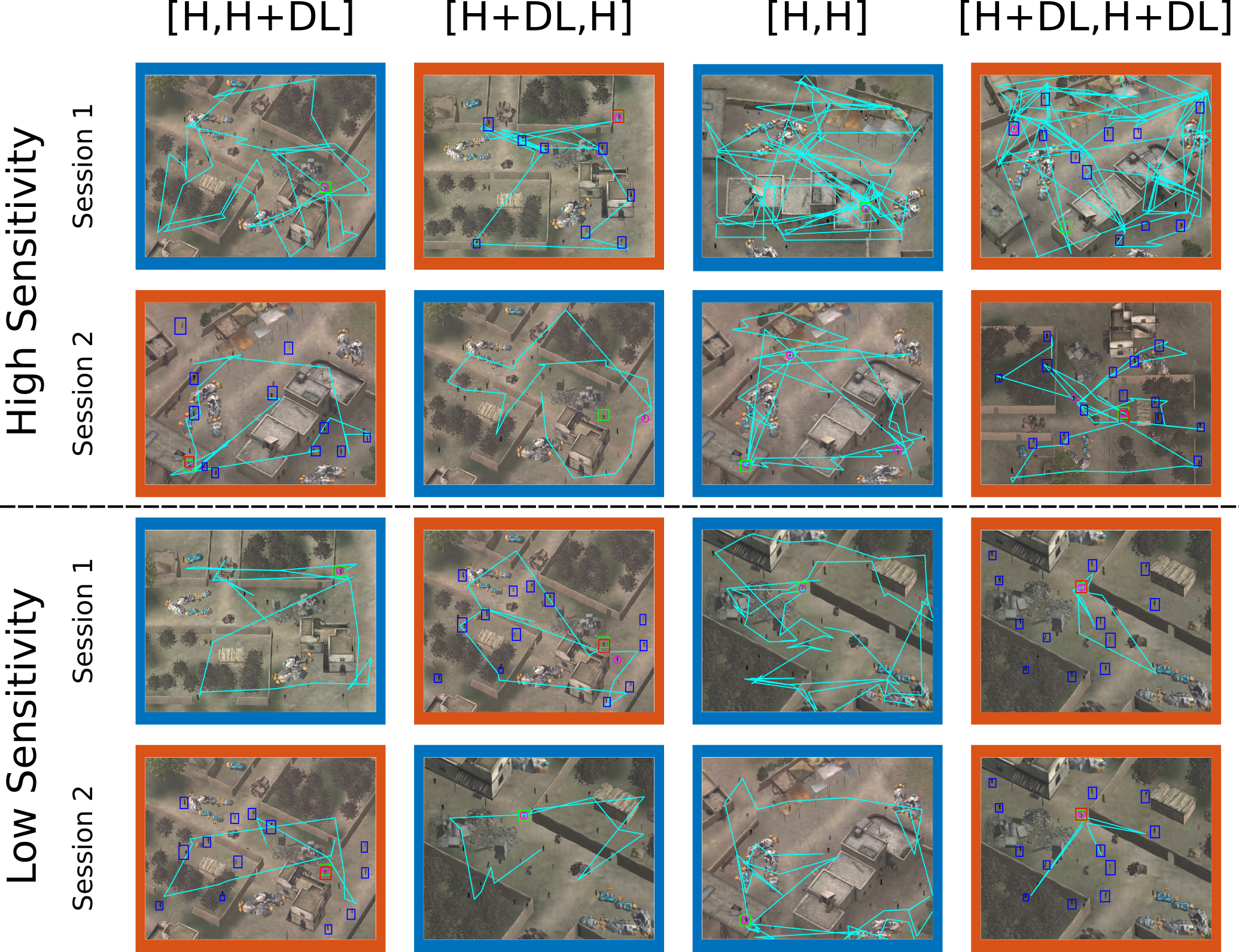}
\caption{Visualization of how visual search strategies change when the DL system is on across all conditions.
The lines in cyan represent the saccadic trajectories starting from the center. 
Boxes in blue are the DL system's detection for friend, and boxes in red are detections
for targets. The box in green shows the ground truth location of the target, and circles in magenta represent the human observer's click (localization).
All stimuli in this plot only have one target. Figure better viewed when zoomed in.}
\label{fig:Saccadic_Plots}
\vspace{-10pt}
\end{figure}

Performance metrics may change as a function of the DL system as well as over each session, but how will human behaviour change
as a function of such conditions? In this subsection we decided to investigate the role of eye-movements in decision making and how 
they may be related to performance levels. More specifically we computed the euclidean distance in degrees of visual angle between the observer's fixation location $f$ and the closest of all possible targets $\bar{t}$ as shown in Eq.~\ref{eq:dist_fix}: 
%These distances
%are shown in the boxplots of Figure~\ref{fig:FixationDistance}.
\begin{equation}
\label{eq:dist_fix}
 D(f,\bar{t}) = \min(\bigcup_i ||f-t_i|| )
\end{equation}
To investigate such question, we decided
to create boxplots of the first 5 fixations across all observers
split in each one of the viewing conditions and also by sensitivity. This can be seen in Figure~\ref{fig:FixationDistance} which suggests
that generally, observers who are enhanced when the DL system is on, 
fixate at a target (contingent
to a target being present) by the third fixation. 
Thus we see how the DL system enhances fixating at the target with fewer eye movements. Qualitative and complimentary plots to this can be observed in Figure~\ref{fig:Saccadic_Plots}, where we show sample gaze and scan path of observers when performing search in all of these conditions. 

What is most revealing about the homogeneity in fixating first at a target with the DL system on, is that this result might explain how most observers either from the high or low sensitivity group may achieve a boost in target detectability $d'$ as shown previously in Table~\ref{table:SCORCH_Summary}.

\section{Main Takeaways from Analysis}
\begin{enumerate}
 \item Target detection performance: The DL system reduces the False Alarm rate per Image on average across observer groups of both high/low 
 sensitivity.
 \item Receiving Operator Characteristics: We found an interaction where only the
 human observers with high sensitivity perform better than the DL system, while the low sensitivity group does not surpass individual DL system performance, even when aided with the DL system itself.
 \item Viewing time and number of trials: The Deep Learning system only increases the number of trials for the low sensitivity group.
 \item Pattern of eye movements: The DL system encourages fixating at the target by the 3rd fixation, independent of other factors.
\end{enumerate}

\section{Discussion}
While there has been a great maturation in terms of success of deep learning systems regarding object detection, 
there are still many limitations in object detection, such as: adversarial 
examples~\cite{goodfellow2014explaining}, fine-grained detection~\cite{hariharan2017object}, small objects(targets)~\cite{eggert2017improving}.
Adversarial examples have clearly exposed important limitations in current deep learning systems, and while having an experimental setup of 
visual search with and without adversarial 
examples would be interesting, it is not the focus of our work.
The outcome is somewhat predictable and guaranteed: humans would achieve a higher recognition rate than computers 
-- yet we do not
discard the possibility that performing a study similar to ours with the presence of adversarial images is 
relevant and should be explored in future work. 
On the other hand, future work regarding integrating
human and machines in visual search in the presence of \textit{human-like} adversarial examples~\cite{elsayed2018adversarial} might also be of great interest as explored in  the recent work of 
Finlayson~\textit{et~al.}~\cite{finlayson2018adversarial} applied to medical images.

In this paper, we thus centered our efforts in studying a more real and applicable problem which is fine-grained small object detection 
and classification with a limited number of 
training exemplars
that uses a commonly deployed pre-trained VGG16~\cite{simonyan2014very}. 
We found that, for a current DL system, its influence on human search performance interacts with the observers' sensitivity. This highlights the complexity of integration of DL systems with humans experts.  It is likely that these interactions also depends on the performance level of the DL system as well as the observers' trust on the DL system.   

With the recent surge of DL systems applied to Medical imaging, we believe that these experimental insights will 
be transferable to such and other human-machine collaborative domains.

\vspace{-10pt}
\subsubsection*{Acknowledgments}
\vspace{-5pt}
This work was supported by the Institute for Collaborative Biotechnologies
through contract W911NF-09-0001 with the U.S. Army Research Office.
\newpage

{\footnotesize
\bibliographystyle{ieee}
\bibliography{egbib}
}

\end{document}